\documentclass[conference]{IEEEtran}
\IEEEoverridecommandlockouts
\usepackage{cite}
\usepackage{amsmath,amssymb,amsfonts}
\usepackage{algorithmic}
\usepackage{graphicx}
\usepackage{textcomp}
\usepackage{xcolor}
\usepackage{orcidlink}
\usepackage{booktabs}
\usepackage{multirow}
\usepackage{threeparttable}
\def\BibTeX{{\rm B\kern-.05em{\sc i\kern-.025em b}\kern-.08em
    T\kern-.1667em\lower.7ex\hbox{E}\kern-.125emX}}
\usepackage{mathtools}
\DeclarePairedDelimiterX{\infdivx}[2]{(}{)}{%
  #1\;\delimsize\|\;#2%
}

\usepackage{hyperref}

\begin{document}

\title{SlimSeiz: Efficient Channel-Adaptive Seizure Prediction Using a Mamba-Enhanced Network
\\
}

\author{ 
Guorui~Lu\orcidlink{0009-0005-3455-0670},
Jing~Peng,
Bingyuan~Huang,
Chang~Gao\orcidlink{0000-0002-3284-4078},
Todor~Stefanov\orcidlink{0000-0001-6006-9366},
Yong Hao,
Qinyu~Chen\orcidlink{0009-0005-9480-6164}

\thanks{Corresponding authors: Guorui Lu (g.lu@liacs.leidenuniv.nl), Qinyu Chen (q.chen@liacs.leidenuniv.nl) and Hao Yong (haoyong@shsmu.edu.cn).}
\thanks{Guorui Lu, Todor Stefanov, and Qinyu Chen are with the Leiden Institute of Advanced Computer Science (LIACS), Leiden University, The Netherlands.}
\thanks{Jing Peng, Bingyuan Huang, and Hao Yong are with Renji Hospital, School of Medicine, Shanghai Jiao Tong University, Shanghai, China.}
\thanks{Chang Gao is with the Department of Microelectronics, Delft University of Technology, The Netherlands.}
}

\maketitle

\begin{abstract}
Epileptic seizures cause abnormal brain activity, and their unpredictability can lead to accidents, underscoring the need for long-term seizure prediction. Although seizures can be predicted by analyzing electroencephalogram (EEG) signals, existing methods often require too many electrode channels or larger models, limiting mobile usability.
This paper introduces a SlimSeiz framework that utilizes adaptive channel selection with a lightweight neural network model. SlimSeiz operates in two states: the first stage selects the optimal channel set for seizure prediction using machine learning algorithms, and the second stage employs a lightweight neural network based on convolution and Mamba for prediction. On the Children’s Hospital Boston-MIT (CHB-MIT) EEG dataset, SlimSeiz can reduce channels from 22 to 8 while claiming a satisfactory result of 94.8\% accuracy, 95.5\% sensitivity, and 94.0\% specificity with only 21.2\,K model parameters, matching or outperforming larger models' performance. 
We also validate SlimSeiz on a new EEG dataset, SRH-LEI, collected from Shanghai Renji Hospital, demonstrating its effectiveness across different patients.
The code and SRH-LEI dataset are available at~\href{https://github.com/guoruilu/SlimSeiz}{https://github.com/guoruilu/SlimSeiz}.
\end{abstract}

\begin{IEEEkeywords}
Deep learning, seizure prediction, state-space model, convolutional neural network, healthcare.
\end{IEEEkeywords}

\section{Introduction}
\begin{figure}[t!]
  \centering
  \includegraphics[width=0.85\columnwidth]{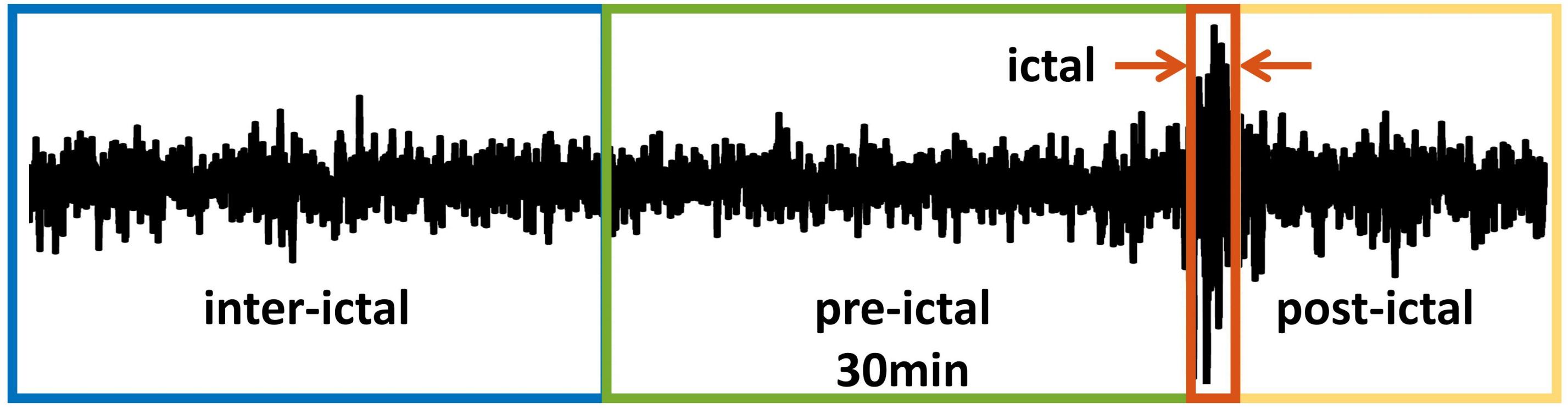}
  \vspace{-0.3cm}
  \caption{A segment of a patient’s scalp EEG from the CHB-MIT dataset. The states can be divided into inter-ictal, pre-ictal, ictal, and post-ictal.}
  \label{fig:seizure_states}
\end{figure}

Epileptic seizure affects more than 50 million patients around the world according to the World Health Organization (WHO)~\cite{world2006neurological}. Its accompanying symptoms can cause sudden and unforeseen accidents, making patients vulnerable to injury~\cite{zhou2018epileptic}. Therefore, continuous patient monitoring and accurate seizure prediction are essential for improving safety and quality of life.
Electroencephalogram (EEG), an electrical recording of brain activity, is a key diagnostic tool for clinicians assessing epilepsy. Fig.~\ref{fig:seizure_states} illustrates that epilepsy EEG signals can be classified into four states: inter-ictal, pre-ictal, ictal, and post-ictal. Seizure prediction aims at identifying the pre-ictal state. However, due to the complexity of EEG signals and the prolonged nature of seizures, manual identification is impractical, requiring algorithms for automatic identification.

During the past decade, many deep learning techniques have been explored to build seizure prediction systems \cite{abdelhameed2021efficient, daoud2019efficient, chen2024epilepsy,lu2023epileptic, tian2021new, ra2021novel, affes2022personalized,jana2021deep, jana2023efficient}. However, most existing methods rely on more than 20-electrode EEG channels to achieve this~\cite{abdelhameed2021efficient, daoud2019efficient, chen2024epilepsy, lu2023epileptic, tian2021new}. Although effective, using over 20 channels limits patient mobility, increases discomfort, and complicates hardware device design, making these methods unsuitable for mobile and long-term monitoring.
Several studies~\cite{ra2021novel, affes2022personalized, jana2021deep, jana2023efficient} have explored using fewer channels for seizure prediction. However, some of these works~\cite{ra2021novel, affes2022personalized} have struggled to achieve 80\% prediction accuracy. While other methods~\cite{jana2021deep, jana2023efficient} have surpassed 90\% prediction accuracy using neural network models, they require over 100\,K model parameters, making such large models less suitable for long-term use on mobile devices.

To address the aforementioned issues, we propose a framework, called \textbf{SlimSeiz}, for efficient seizure prediction using a lightweight Mamba-enhanced neural network with adaptive EEG channel selection.
Our main novel contributions are:
\begin{enumerate}
\item A machine learning-based adaptive channel selection method that reduces the number of EEG channels from 22 to 8 while maintaining satisfactory accuracy.
\item A lightweight neural network model with only 21.2\,K parameters, consisting of one-dimensional convolutions and the Mamba~\cite{gu2023mamba} block.
\item  new dataset called the Shanghai Renji Hospital-LEI (SRH-LEI) EEG dataset.  
\item An experimental evaluation of SlimSeiz on the Children’s Hospital Boston-MIT (CHB-MIT) EEG dataset and the SRH-LEI EEG dataset. SlimSeiz achieves 94.8\% accuracy, 95.5\% sensitivity, and 94.0\% specificity on the CHB-MIT dataset, and 92.7\% accuracy, 94.7\% sensitivity, and 90.7\% specificity on the SRH-LEI dataset.
\end{enumerate}

\section{The SlimSeiz Framework}
In this section, we start by explaining how the EEG data is pre-processed (Section \ref{sec:pre-processing}). Then, we describe the two main parts of SlimSeiz: the channel selection method (Section~\ref{sec:ch_sel}) and the network model (Section~\ref{sec:network}). The channel selection module picks the top $k$ channels that are most important for predictions. The neural network model then makes predictions based on these selected channels.

\begin{figure*}[t!]
  \centering
  \includegraphics[width=0.95\textwidth]{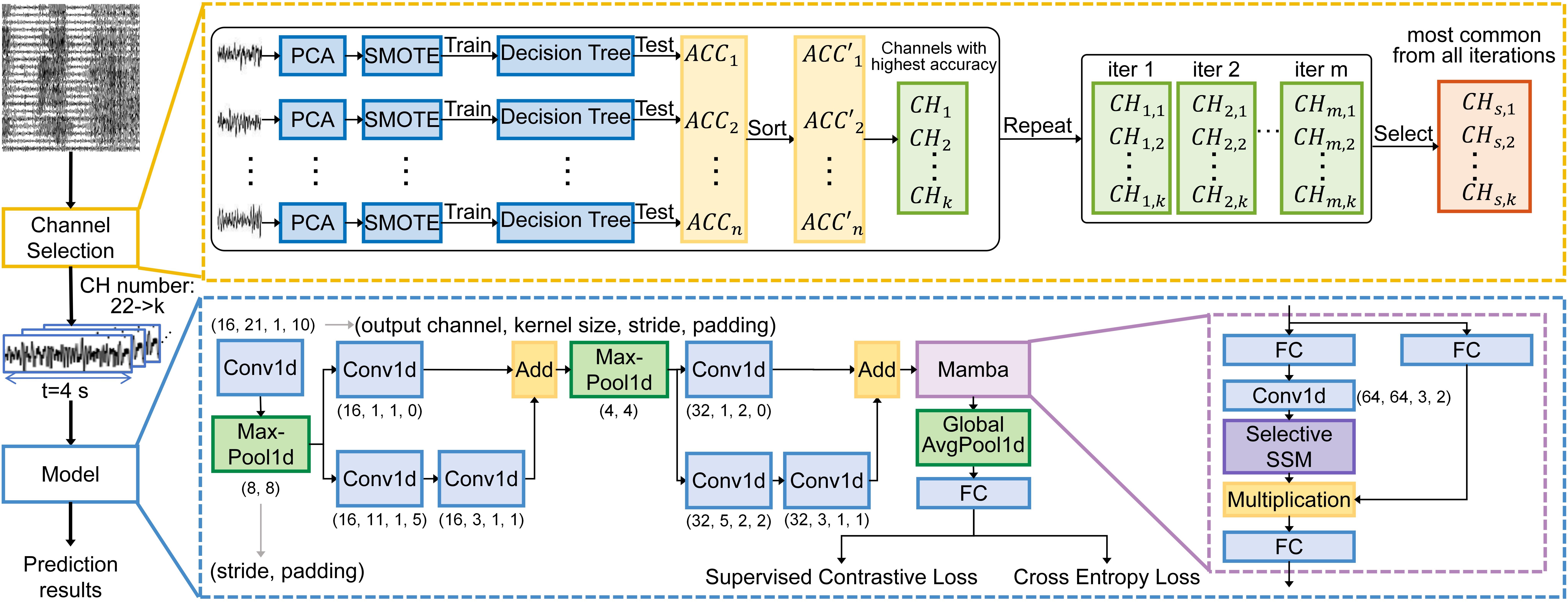}
  \vspace{-0.35cm}
  \caption{The overview of SlimSeiz framework. Data from the original 22 channels are input into the channel selection module to select the $k$ (an adjustable hyperparameter) channels that contribute most significantly to seizure prediction, using accuracy (ACC) as the metric. The selected $k$ channels are then sent to the neural network for training and validation. All activation functions are not shown in the figure. The convolutional module in the Mamba block uses the SiLU function, while the other convolution modules use the ReLU function. The FC layers at the input of the Mamba block increase the channel count from 32 to 64, and the output FC layer reduces it from 64 to 32.}
  \label{fig:selection_and_model}
\end{figure*}

\subsection{EEG Data Pre-processing}
\label{sec:pre-processing}

In this work, we use the CHB-MIT and the newly collected SRH-LEI EEG datasets (details are described in Section~\ref{sec:datasets}).
Due to the imbalanced nature of the recorded EEG data, where the duration of seizure events is significantly shorter than inter-ictal and post-ictal events, necessary data pre-processing operations can have a considerable impact on the model training and its predictive accuracy. Some previous work~\cite{abdelhameed2021efficient, jana2023efficient, jana2021deep, zhao2022patient} discarded the majority of the data to achieve a balanced dataset. In this work, we choose to use overlapping sliding windows with adjustable step sizes to increase the number of pre-ictal segments, similar to the approach used in~\cite{chen2024epilepsy}.

The goal of seizure prediction is to identify pre-ictal states. Therefore, ictal segments are discarded because the seizure has already begun. The remaining segments, except pre-ictal, are grouped into a single category for simplicity. We set the pre-ictal segment length to 30 minutes, as prior work indicates this period contains important EEG information for pre-ictal/inter-ictal recognition~\cite{toraman2021automatic}. If the gap between two seizures is less than 30 minutes, we combine them due to difficulties in distinguishing post-ictal and pre-ictal states. Each EEG recording is split into 4-second segments given as input to the neural network model within SlimSeiz.

\subsection{Channel Selection Method}
\label{sec:ch_sel}
Long-term monitoring requires selecting a smaller subset of channels to reduce the computational load on the seizure predicting device and to enhance the patient mobility and comfort. However, the challenge lies in identifying which channels contribute most to the seizure prediction. In our work, we use the prediction accuracy achieved by each individual channel as an indicator of its contribution, based on the assumption that channels with higher prediction accuracy contain more useful information. Although the channel selection is a one-time effort for each patient, we aim to minimize the channel selection time as much as possible. Thus, we propose a channel selection method based on traditional machine learning that can yield results more quickly compared to neural networks.

The channel selection method is illustrated in the top part of Fig.~\ref{fig:selection_and_model} as a workflow. First, we split each channel data into training and testing segments. Given the fact that traditional machine learning (ML) methods have limited feature extraction capabilities, we use longer 5-second data segments to provide the ML model with more information. Then, we apply Principal Component Analysis (PCA)~\cite{abdi2010principal} to reduce the dimension for the training and testing segments separately while preserving key information, thereby reducing subsequent computational overhead. The Synthetic Minority Over-sampling Technique (SMOTE)~\cite{chawla2002smote} is employed to balance the dimension-reduced training set, which occupies less memory compared to using overlapping sliding windows. 
Next, we train and test a decision tree on the processed dataset to obtain the prediction accuracy for the current channel. 
By executing the aforementioned procedure for each channel, we generate a list of accuracy numbers and select the top $k$ channels based on these accuracy numbers to form a channel subset. To mitigate the influence of random factors, such as the channel data splitting into training and testing segments, on the results, we repeat this procedure $m$ times to obtain $m$ channel subsets. 

Finally, we count the occurrences of each channel across all subsets and select the $k$ channels that appear most frequently as the final result. The subset size $k$ and the number of iterations $m$ are two hyperparameters. In this study, we set $m$ to 30 to balance the channel selection time and elimination of random factors. The optimal value of $k$ is determined through relevant experiments to achieve a balance between reducing the number of channels and maintaining high accuracy of the neural network model we have devised to predict seizures.

\subsection{Neural Network Model Architecture}\label{sec:network}
As shown in the bottom part of Fig.~\ref{fig:selection_and_model}, our lightweight model for seizure prediction consists of convolutional feature extraction, Mamba, and classification head.

\subsubsection{Convolutional Feature Extraction}
We implement the feature extraction from the EEG input data by stacking one-dimensional (1D) convolutional layers that form the main body of our neural network model for seizure prediction. As shown in Fig.~\ref{fig:selection_and_model}, a larger convolution kernel of size 21 is applied on the EEG input data in the first convolutional layer. This is followed by two residual connections, each consisting of three convolutional layers with different smaller kernel sizes. Different kernel sizes help the network extract features at various temporal resolutions. Max pooling and global average pooling are employed to reduce the feature dimensions and computational load. It is worth mentioning that previous work~\cite{zhao2022patient, chen2024epilepsy, song2022eeg, dissanayake2021deep, zhao2021energy} primarily uses over 20 channels for seizure prediction, treating channels and time as the height and width of an image, respectively, and thus tend to use 2D convolutional neural network to extract spatial and temporal features. In our work, however, we use fewer channels, thereby reducing the complexity of spatial features and eliminating the need for parameter- and compute-intensive 2D convolutional networks.


\subsubsection{Mamba}
The Mamba block is based on the State Space Model (SSM). Its structure, as shown in Fig.~\ref{fig:selection_and_model}, uses a fully connected (FC) layer at the input to increase the channel count for capturing more complex features. The subsequent convolutional layer and SSM establish short-term and long-term sequential dependencies, respectively. The FC layer at the output reduces the channel count to maintain consistent input and output dimensions. This structure enables the Mamba block to efficiently capture dependencies within the sequence~\cite{qu2024survey}. Mamba is proposed as an alternative to Transformer to reduce the computational load in the multi-head self-attention mechanism~\cite{zhang2024mamba} and has shown promising potential in processing physical signals, such as audio signals~\cite{chen2024rawbmamba, lin2024audio}.  

Mamba's abilities to capture dependencies in sequential data and reduce computation complexity match our goal to design a lightweight model for processing of EEG signals. To enhance the model's ability to capture temporal dependencies without increasing the model parameters, we add the Mamba block after the convolutional layers. This choice is due to the limited semantic information of sensor signals at a single time point, making it difficult to establish correlations~\cite{zhang2022if}. Therefore, extracting features through convolutional layers and increasing channel count enriches the semantic information at each time point, allowing the Mamba block to work more effectively.

\subsubsection{Classification Head}
A fully connected layer is used for classification. Cross-entropy loss and supervised contrastive loss are used to supervise the model's training.

\begin{figure*}[htbp]
  \centering
  \includegraphics[width=0.92\textwidth]{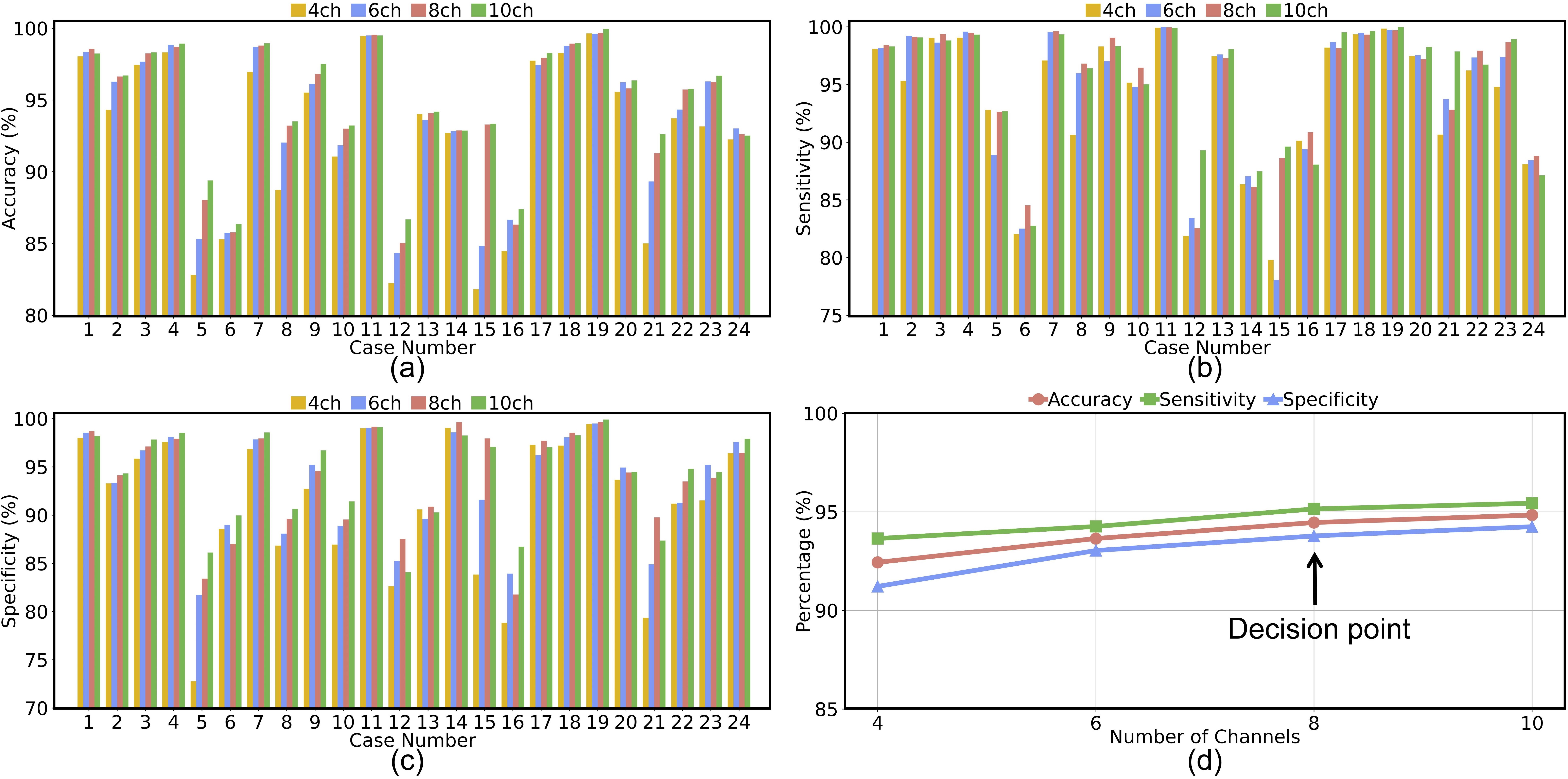}
  \vspace{-0.4cm}
  \caption{Performance: (a) accuracy (b) sensitivity (c) specificity across different cases for various channel counts on the CHB-MIT dataset. (d) Overall accuracy, sensitivity, and specificity trends as the number of channels increases from 4 to 10 on the CHB-MIT dataset.}
  \label{fig:ch_results}
\end{figure*}



\section{Experimental Evaluation and  Results}
In this section, we evaluate our proposed SlimSeiz framework to demonstrate its potential and efficiency. 

\subsection{Experimental Setup and Datasets}\label{sec:datasets}
We evaluate SlimSeiz using two datasets.
First, we utilize the CHB-MIT scalp long-term EEG dataset. This dataset comprises recordings from 23 children with intractable seizures at the Children’s Hospital Boston, resulting in 24 cases (with two cases derived from a single patient). The duration of EEG recordings for each patient ranges from 19 to 165 hours, totaling approximately 983 hours. The sampling frequency for signal collection is 256 Hz.
Second, we introduce and utilize the new SRH-LEI EEG dataset, which we have collected at Shanghai Renji Hospital, Shanghai Jiao Tong University. This dataset includes data from 8 patients, consisting of 4 males and 4 females, aged between 14 and 77 years old. The data collection duration ranges from 17 to 24 hours, with a sampling frequency of 128 Hz. Detailed information about the dateset features is provided in Table~\ref{tab:SRHLEI}.

We use accuracy (ACC), sensitivity (SENS), and specificity (SPEC) as evaluation metrics. 
SENS measures the model’s ability to correctly identify actual pre-ictal segments, higher sensitivity means fewer missed pre-ictal cases. SPEC indicates how well the model identifies actual inter-ictal segments, higher specificity reflects fewer false alarms for pre-ictal states.



\begin{table}[t]
\centering
\caption{Details of the SRH-LEI dataset}
\label{tab:SRHLEI}
\vspace{-0.3cm}
\begin{tabular}{ccccc}
\toprule[1pt]
\textbf{Case }& \multirow{2}{*}{\textbf{Gender} } & \multirow{2}{*}{\textbf{\#Seizures}} & \textbf{Seizure duration}& \textbf{EEG duration}\\ 
\textbf{No.}& \multirow{2}{*}{} & \multirow{2}{*}{} & \textbf{mm:ss} & \textbf{hh:mm:ss}\\ \hline
1                 & Female                  & 12                     & 00:19         & 18:29:23           \\
2                 & Female                  & 95                     & 13:53              & 23:59:50      \\
3                 & Male                  & 8                     & 27:57             & 21:33:56       \\
4                 & Female                  & 46                     & 1:12                & 23:58:51    \\
5                 & Male                  & 32                     & 67:10              & 22:30:45      \\
6                 & Male                  & 3                     & 24:7              & 23:58:51      \\
7                 & Male                  & 45                     & 479:12               & 22:47:10     \\
8                 & Female                  & 33                     & 270:34                & 17:08:34    \\
\bottomrule[1pt]
\end{tabular}%
\end{table}
\begin{table*}[t]
\centering
\caption{Performance Comparison of SlimSeiz with other works. The best metric number is \textbf{bolded}; the second best is \underline{underlined}.}
\label{tab:performance_comparison}
\vspace{-0.3cm}
\begin{threeparttable}
\begin{tabular}{cccccccccc}
\toprule[1pt]
\multirow{2}{*}{\textbf{Ref}}                                          & \multirow{2}{*}{\textbf{Cases}} & \multirow{2}{*}{\textbf{Model}} & \multirow{2}{*}{\textbf{Channels}} & \multirow{2}{*}{\textbf{Parameter}} & \textbf{Feature } & \multirow{2}{*}{\textbf{Validation}} & \multicolumn{3}{c}{\textbf{Performance}} \\ \cline{8-10} 
&                                 &                                 &                                    &                                     &                                     \textbf{ Extraction}        &                                      & ACC          & SENS        & SPEC        \\ \hline
Zhang et all. 2020 \cite{zhang2019epilepsy}           & 15                              & CNN                             & \underline{18}                                 & 194.6K                              & CSP                                          & LOOCV                                & 90.0         & 92.2        & 92.0        \\
Buyukccakir et al. 2020 \cite{buyukccakir2020hilbert} & 10                              & MLP                             & \underline{18}                                 & 40.9K                               & HVD                                          & 10-Fold CV                           & -            & 89.9        & -           \\
Baghdadi et al. 2020 \cite{baghdadi2020robust}        & 24                              & Deep LSTM                       & \underline{18}                                 & \textgreater{}3M                    & Raw EEG                                      & 10-Fold CV                           & 88.9         & 84.0        & 90.0        \\
Tian et al. 2021 \cite{tian2021new}                   & 7                               & Spiking CNN                     & 23                                 & \textbf{10.3K}                               & Raw EEG                                      & 80-20 split                          & -            & 95.1        & \textbf{99.2}        \\
Zhao et al. 2022 \cite{zhao2022patient}               & 19                              & AddNet-SCL                      & 22                                 & 120K                                & Raw EEG                                      & LOOCV                                & -            & 94.9        & -           \\
Lu et al. 2023 \cite{lu2023epileptic}                 & 11                              & CBAM-3D CNN-LSTM                & 22                                 & -                                   & STFT                                         & LOOCV                                & \textbf{97.9}         & \textbf{98.4}        & -           \\
Chen et al. 2024 \cite{chen2024epilepsy}              & 24                              & Spiking Conformer               & 22                                 & 40.3K                               & Raw EEG                                      & 10-Fold CV                           & 93.1         & \underline{96.8}        & 89.5        \\
\textbf{Our work}                                                      & 24                              & Convolution Mamba               & \textbf{8}                                 & \underline{21.2K}                               & Raw EEG                                      & 10-Fold CV                           & \underline{94.8}         & 95.5        & \underline{94.0}        \\ 
\bottomrule[1pt]
\end{tabular}%
\end{threeparttable}
\end{table*}

\begin{table}[t]
\centering
\caption{Performance of SlimSeiz on the SRH-LEI dataset}
\label{tab:SJTU_res}
\vspace{-0.3cm}
\begin{tabular}{cccc}
\toprule[1pt]
\textbf{Case No.} & \textbf{Accuracy (\%)} & \textbf{Sensitivity (\%)} & \textbf{Specificity (\%)} \\ \hline
1                 & 93.7                  & 96.1                     & 91.2                     \\
2                 & 74.3                  & 77.6                     & 71.0                     \\
3                 & 98.0                  & 99.0                     & 97.0                     \\
4                 & 94.5                  & 95.6                     & 93.7                     \\
5                 & 96.0                  & 98.8                     & 93.3                     \\
6                 & 99.0                  & 99.5                     & 98.4                     \\
7                 & 90.5                  & 92.3                     & 88.6                     \\
8                 & 95.6                  & 99.1                     & 92.2                     \\
\hline
\textbf{Average}                 & 92.7                  & 94.7                     & 90.7                     \\ 
\bottomrule[1pt]
\end{tabular}%
\end{table}
\subsection{Optimal Number of Channels}\label{sec:channel_number_experiment}
To determine the optimal channel counts, we evaluated the model's performance using channel counts of 4, 6, 8, and 10. In this experiment, we randomly split the data into 80\% for training and 20\% for testing to assess the effectiveness of each channel configuration.

As shown in Fig.~\ref{fig:ch_results}, our experimental results indicate that increasing the number of channels beyond 8 does not lead to significant performance improvements. More importantly, for most cases (patients), an accuracy of over 90\% can be achieved using just 4 channels, and for some patients (e.g., case 01 and 07), the accuracy even exceeds 95\%. Therefore, increasing the number of channels offers little benefit for most patients. Only for a few patients (e.g., case 15 and 21) with an accuracy below 90\% using 4 channels, increasing the number of channels results in a noticeable performance enhancement (up to 10\%). Considering both performance gains and the associated complexity of the device with the number of channels, we conclude that 8 is the optimal number of channels and we use this result in subsequent experiments.

\subsection{Performance and Comparison Results}\label{sec:comparison_res}
To validate the effectiveness of the Mamba block, we replace it with a convolutional block of a similar parameter scale.
The total parameters for the network with the Mamba block and the network with the replaced convolutional block are 21.2K and 22.2K, respectively. While having a slightly smaller model size, the network with the Mamba block achieves an average accuracy that is 0.1\% higher than the network with the replaced convolutional block in 10-fold cross validation (CV), indicating the effectiveness of the Mamba block in lightweight seizure prediction neural networks.

We also conduct experiments on the SRH-LEI EEG dataset. The results when utilizing 8 selected channels are shown in Table~\ref{tab:SJTU_res}. SlimSeiz achieves an average of over 90\% for each metric, demonstrating its generalization on adult patients.

Finally, we compare our framework with state-of-the-art seizure prediction methods on the CHB-MIT dataset. 
As shown in Table~\ref{tab:performance_comparison}, our framework using only 8 channels and a small model (21.2K parameters) achieves comparable average accuracy to methods using over 18 channels and larger models, highlighting its suitability for long-term wearable monitoring.

\section{Conclusion and Future Work}

We present SlimSeiz, a lightweight framework for efficient seizure prediction, using only 21.2\,K model parameters and reducing EEG channels from 22 to 8 while matching state-of-the-art performance on the CHB-MIT dataset. Additionally, SlimSeiz is validated on the newly collected SRH-LEI EEG dataset, which will be released soon. We found that Channels P3-O1, P8-O2, C3-P3, C4-P4, FZ-CZ, and P4-O2 are selected in over 17 patients, paving the way for future patient-independent systems with fewer channels. 



\bibliographystyle{IEEEtran}
\bibliography{refs.bib}
\end{document}